\pdfoutput=1

\documentclass[11pt]{article}

\usepackage{EMNLP2022}

\usepackage{times}
\usepackage{latexsym}

\usepackage[T1]{fontenc}

\usepackage[utf8]{inputenc}

\usepackage{microtype}

\usepackage{inconsolata}

\usepackage{algorithm}
\usepackage{algorithmic}
\usepackage{booktabs} 
\usepackage{enumitem}
\usepackage{bbm}
\usepackage{csquotes}
\usepackage{multirow}
\usepackage{multicol}
\usepackage{comment}
\usepackage{xcolor}

\usepackage{todonotes} 

\newcommand{\ar}[1]{\textcolor{red}{\bf\small [#1 --AR]}}

\definecolor{forestgreenweb}{rgb}{0.13, 0.55, 0.13}
\definecolor{cadmiumgreen}{rgb}{0.0, 0.42, 0.24}
\definecolor{crimsonglory}{rgb}{0.75, 0.0, 0.2}
\newcommand{\maxle}[1]{\textcolor{crimsonglory}{\bf\small [#1 --Max]}}

\newcommand{\mh}{\hat{m}}

\definecolor{anaphor}{HTML}{EF8810}
\definecolor{gold_antecedents}{HTML}{1F72D4}
\definecolor{wrong_antecedents}{HTML}{FF0000}

\newcommand\mice{\textsc{Mice}}
\newcommand\micesamp{\textsc{Mice-Sampling}}

\newcommand\kate{\textsc{Kate}}
\newcommand\product{\textsc{Product}}
\newcommand\etoe{E2E}

\newcommand\chemuref{\textsc{ChEMU-Ref}}
\newcommand\procbert{\textsc{ProcBERT}}

\title{Few-Shot Anaphora Resolution in Scientific Protocols via \\ Mixtures of In-Context Experts}

\author{Nghia T. Le, Fan Bai, Alan Ritter \\
  School of Interactive Computing \\
  Georgia Institute of Technology \\
 \texttt{\{nle18,fan.bai,alan.ritter\}@cc.gatech.edu} 
\\}

\begin{document}
\maketitle

\begin{abstract}
Anaphora resolution is an important task, which traditionally has required costly supervised training datasets for each new language, text genre, and domain.
Meanwhile, prompting large language models with a few in-context examples has emerged as a promising approach to reduce labeling costs, however there are a number of challenges in applying in-context learning to resolve anaphora. 
In this paper, we present {\sc Mice} (Mixtures of In-Context Experts), which we demonstrate is effective for few-shot anaphora resolution in the domain of scientific protocols \citep{tamari-etal-2021-process}.  Given only a handful of training examples, {\sc Mice} combines the predictions of hundreds of in-context experts, yielding a 30\% increase in F$_1$ score over a competitive prompt retrieval baseline.  Furthermore, we show {\sc Mice} can be used to train compact student models without sacrificing performance. As far as we are aware, this is the first work to present experimental results demonstrating the effectiveness of in-context learning on the task of few-shot anaphora resolution in scientific protocols.\footnote{Our code and datasets are available at \url{https://github.com/nle18/mice}}

\end{abstract}

\section{Introduction}
\label{sec:intro}

\begin{figure*}[!ht]
    \centering
    \includegraphics[width=\textwidth]{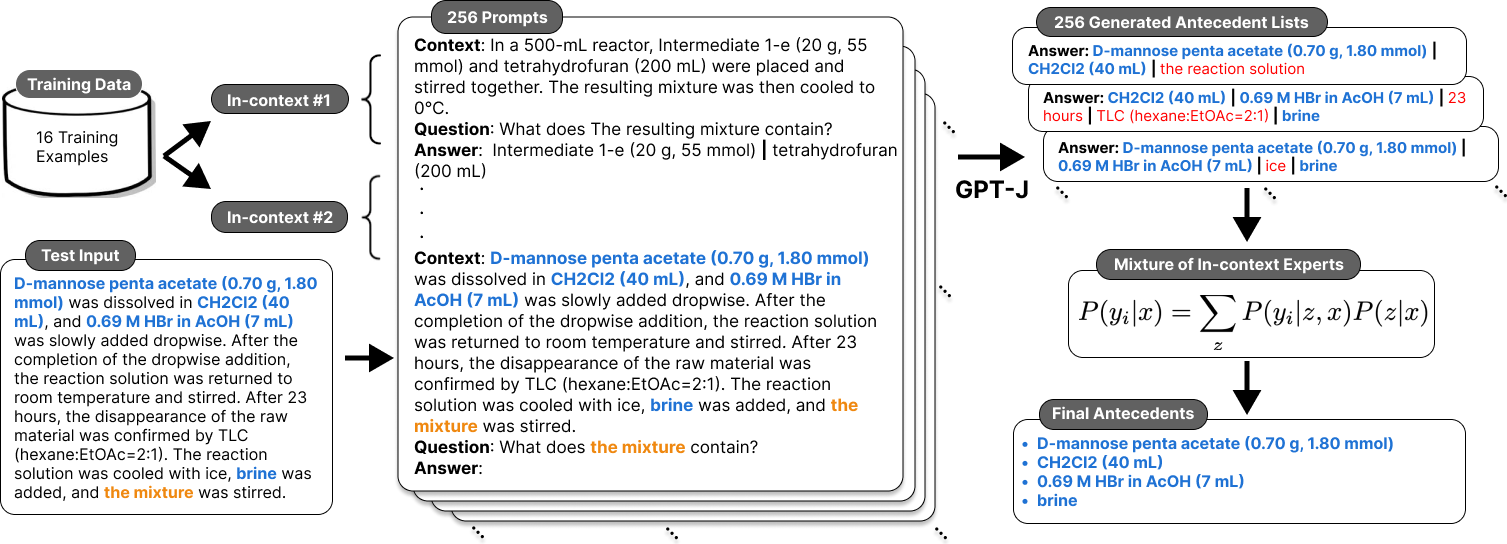}
    \caption{Resolving antecedents of {\em ``the mixture''} in a chemical synthesis procedure using {\sc Mice}. Given a small training set of 16 examples and a test input, we construct 256 prompts, each with two in-context demonstrations. The prompts are then fed into a pre-trained language model (e.g. GPT-J) to generate candidate antecedents. The probabilities of each candidate antecedent are computed and combined in a mixture of in-context experts using a similarity-based gating function. \mice{} then selects the antecedents with the highest probabilities. In the figure, orange, blue, red denote \textbf{\textcolor{anaphor}{the anaphor}}, \textbf{\textcolor{gold_antecedents}{the true antecedents}}, and \textcolor{wrong_antecedents}{incorrect antecedents}, respectively.}
    \label{fig:approach}
\end{figure*}

Prompting large language models (LMs) with in-context demonstrations has enabled surprisingly effective few-shot learning \citep{gpt3}.  
However, more complex linguistic annotations over paragraph-length inputs, such as anaphora and coreference, have proven challenging \citep{yang2022gpt}.
Prompting language models with demonstrations of anaphora and their corresponding antecedents requires encoding long sequences of tokens, limiting the number of demonstrations that can be used within a single prompt.  Furthermore, the performance of in-context learning has been shown to be sensitive to the choice of demonstrations \citep{liu-etal-2022-makes} and their ordering in the prompt \citep{lu-etal-2022-fantastically}.   


To address these challenges, we present \textbf{M}ixtures of \textbf{I}n-\textbf{C}ontext \textbf{E}xperts ({\sc Mice}). We demonstrate {\sc Mice}'s effectiveness on anaphora resolution in chemical synthesis protocols (see examples in Figure \ref{fig:approach}).  Natural language understanding for protocols makes an attractive use-case for few-shot learning, as experimental procedures contain rich coreference and bridging links. Anaphora in protocols are expressed quite differently from those found in high-resource domains (e.g. newswire), and scientific protocols are not easily amenable to annotation by non-expert crowd workers.

{\sc Mice} works as follows.  Given an anaphor, such as {\em ``the mixture''}, it uses in-context learning to predict a list of  substances contained in the mixture that are referenced earlier in the procedure, for example: {\em ``Bromoacetyl bromide''}, {\em ``compound 54''} and {\em ``water''}.  With only a handful of training examples (e.g., 16 or 32), {\sc Mice} generates an ensemble of up to $k^d$  in-context experts, each of which consists of a prompt containing $d$ demonstrations chosen from the $k$ available training examples.  
The experts' predictions are then combined in a mixture model, where mixture weights are computed by comparing embeddings of the input to the demonstrations encoded by each in-context expert.

\begin{figure}[!ht]
    \centering
    \includegraphics[scale=0.23] {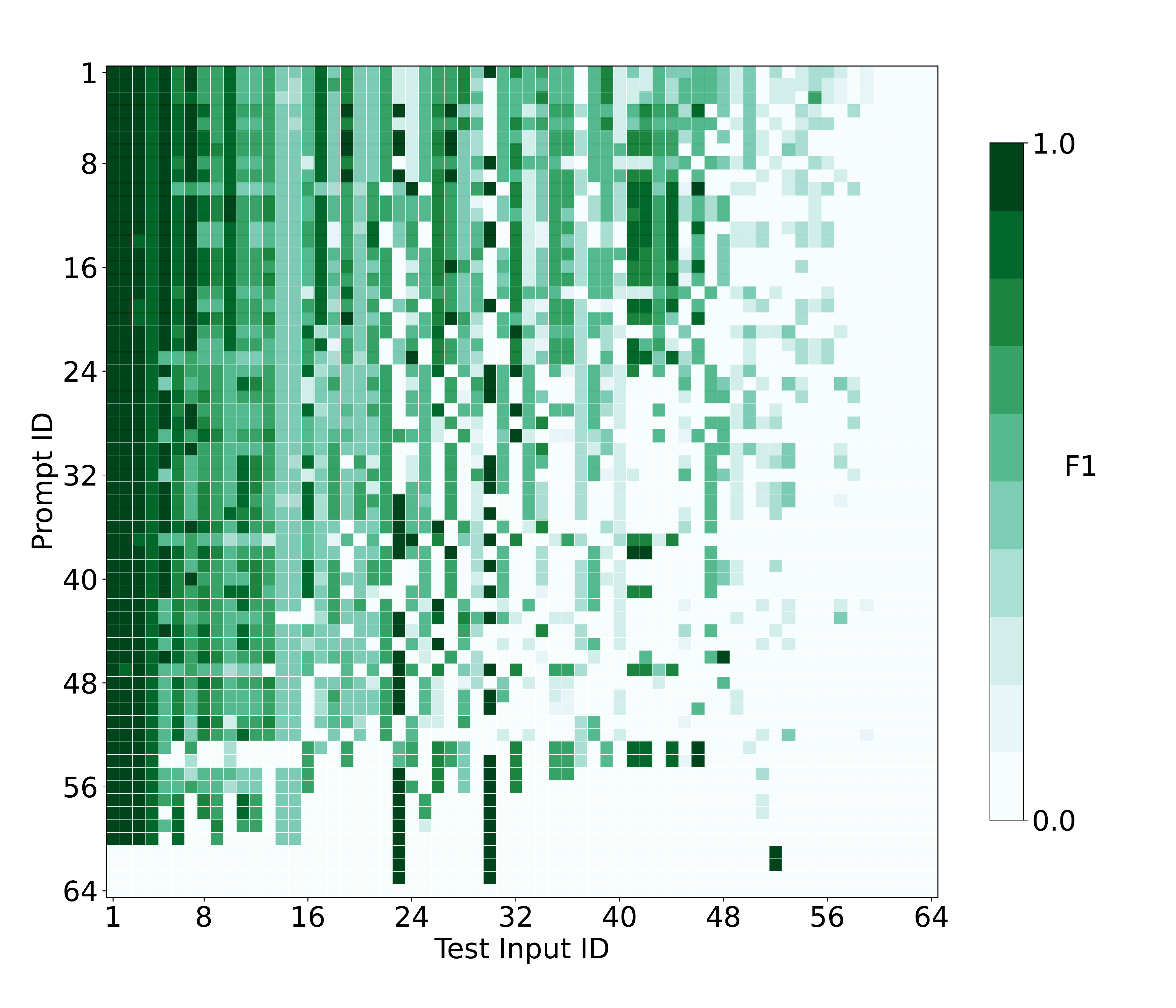}
    \caption{Heatmap visualizing the performance of 64 prompts on 64 sampled anaphors. Each prompt encodes two in-context demonstrations randomly sampled from 8 training examples. Each square represents F$_1$ of a single prompt applied to a single anaphor (typically these are associated with multiple antecedents). The prompts and test inputs are sorted from high (top, left) to low (bottom, right) F$_1$.  Note that no single prompt performs best on all test inputs.  This suggests that it could be beneficial to combine lists of predicted antecedents made independently by many {\em in-context experts}.}
    \label{fig:testID_vs_promptID}
\end{figure}

Although some in-context experts perform better than others, individual prompts act as local experts in different regions of the input space \citep{jacobs1991adaptive, jordan1994hierarchical, SparseMoE}, and no single prompt works better than others on all inputs (see Figure \ref{fig:testID_vs_promptID}).
Furthermore, if the same antecedent is predicted by multiple in-context experts, this provides independent evidence for the prediction, increasing the probability the answer is correct \citep{downeyredundancy}.

In extensive experiments, we show {\sc Mice} significantly improves the performance of in-context learning for anaphora resolution in synthetic procedures.
For example, given 32 demonstrations, a single prompt achieves an F$_1$ score of $38.6$.  By combining the predictions of an ensemble of 256 in-context experts, {\sc Mice} achieves $53.9$ F$_1$. 

While {\sc Mice} consistently improves in-context anaphora resolution, inference in {\sc Mice} is costly, due to the large number of prompts involved.
To address this limitation, we show that fine-tuning compact BERT-based models on data that is automatically labeled by \mice{} can yield performance improvements, while also producing models that support efficient inference \citep{schick2021exploiting,schick-schutze-2021-just,Lang2022Cotraining}.

\section{Anaphora Resolution in Scientific Protocols}
\label{sec:background}

Split-antecedent anaphors \citep{vala-etal-2016-antecedents, yu-etal-2020-free,paun2022scoring} are plural mentions that refer to two or more antecedents in the previous discourse.  For instance in the following text: ``\texttt{[Alice]\textsubscript{antecedent} and [Bob]\textsubscript{antecedent} went to the store. [They]\textsubscript{anaphor} bought some bread.}'' the word ``\texttt{[They]}'' refers to two both ``\texttt{[Alice]}'' and ``\texttt{[Bob]}''.

Similar references to multiple antecedents often appear in chemical synthesis protocols, for example, ``{\em the mixture}''.  These references arise naturally as the result of context change accommodation \citep{webber1992accommodating}, and are crucial for understanding the steps needed to synthesize a molecule \citep{fang-etal-2021-chemu}. Resolving anaphoric references in synthetic protocols could be beneficial for automating protocols described in natural language \citep{sanderson2019automation,vaucher2021inferring}, in addition to automatically extracting chemical reaction databases from scientific literature \citep{lawson2014making,mysore2019materials}.  However, anaphora is costly to annotate \citep{yuan2022adapting} and scientific protocols are not easily amenable to annotation by non-expert crowd workers \citep{kulkarni2018annotated}.  This motivates the need for few-shot learning methods that can resolve anaphora in procedural texts without extensive annotated resources.

\section{Mixtures of In-Context Experts}
\label{sec:in_context_mice}

\begin{figure}[!h]
    \centering
    \includegraphics[width=0.50\textwidth]{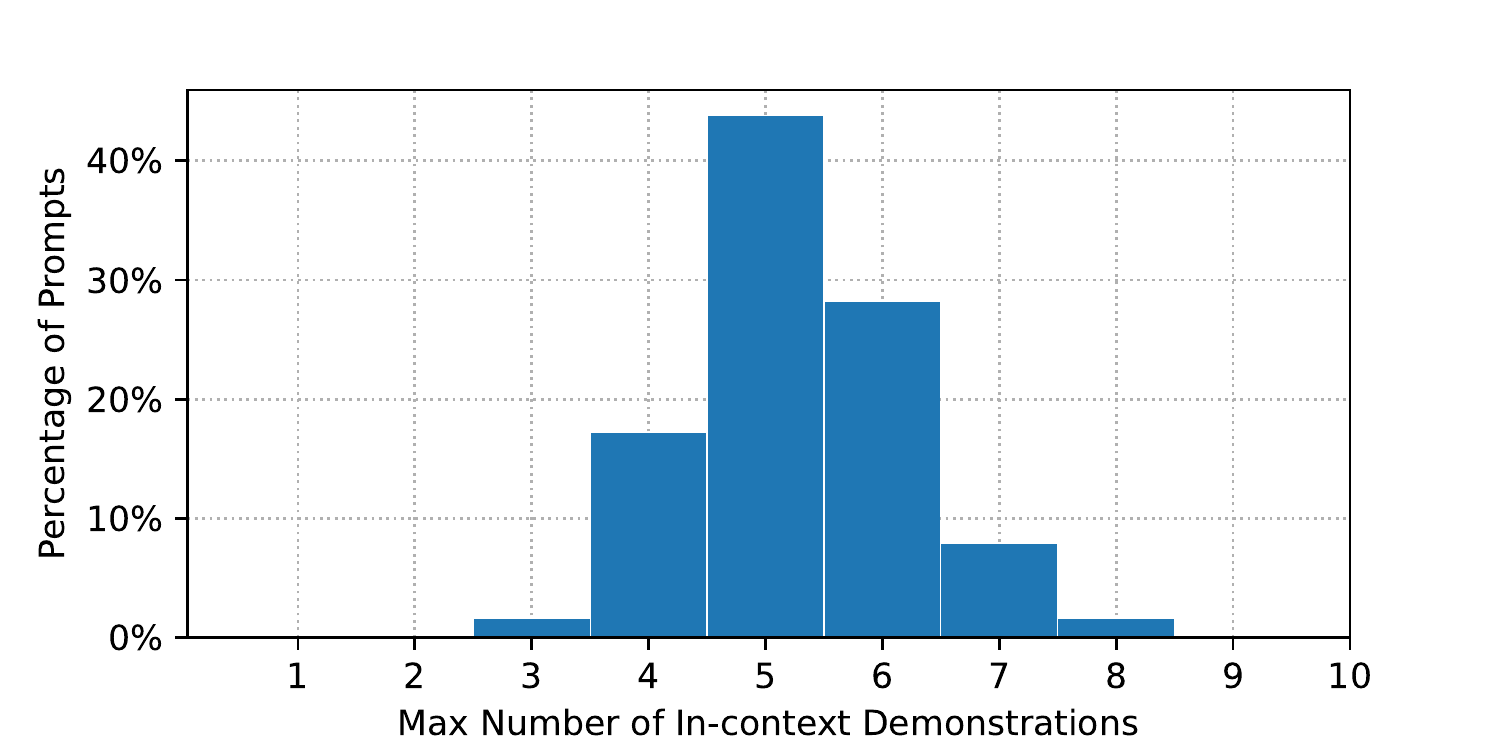}
    \caption{Distribution of the maximum number of in-context demonstrations of an anaphor, synthesis protocol, and corresponding antecedents that can be encoded in a single prompt. We compute the max number of demonstration tokens by subtracting the max sequence length (2048) by a fixed number for generated tokens (256) and the number of tokens for the test input. Demonstrations are randomly sampled until the max number of tokens is reached. Given the longer contexts needed to demonstrate anaphora resolution, a prompt can encode at most 8 demonstrations, much less than the 32  used in \citet{gpt3}.}
    \label{fig:dev64_maxInContextNum}
\end{figure}

While in-context learning has achieved good performance when prompted with a few examples, the performance can vary significantly depending on different prompt design choices \citep{lu-etal-2022-fantastically, liu-etal-2022-makes}. 
Furthermore, anaphora resolution requires paragraph-length contexts, limiting the number of in-context examples that can be encoded in a single prompt (Figure \ref{fig:dev64_maxInContextNum}). We address these challenges with \mice{}. We show \mice{} is an effective method for few-shot anaphora resolution in \S \ref{sec:results}, and demonstrate that it can be used to automatically label data for fine-tuning more compact models, without sacrificing performance, in \S \ref{sec:implementation}.    

\paragraph{In-Context Learning}
We formulate the task of anaphora resolution in synthetic protocols as follows.
The input includes a document $D$ and a query anaphor $a$. Our goal is to identify a set of antecedents $\mathcal{Y} = \{y_0, y_1, ..., y_m\}$ that correspond to text spans in $D$.
To tackle this problem via in-context learning,
we frame it as a SQuAD-style extractive question answering task \citep{wu-etal-2020-corefqa}.
Specifically, as shown in Figure \ref{fig:approach}, each example $(D, a)$ is formatted as the concatenation of document $D$ and template question: \texttt{``What does $a$ contain?''}
An autoregressive language model then completes this sequence by generating $\mathcal{Y}$, with the antecedents separated by a special marker ``{\tt |}''. 
Following the typical approach to in-context learning, the prompt includes a few demonstrations in the prefix and ends with the test input.

\paragraph{Mixture of Experts} For a given test input $x=(D,a)$, we aim to find the antecedents $y_i$ with the highest probabilities $P(y_i|x)$. The notation $y_i$ denotes an antecedent from the union of antecedents generated by all prompts.
\mice{} computes $P(y_i|x)$ using a mixture of experts \citep{jacobs1991adaptive,cho2019mixture}, treating the prompt, $z$, as a latent variable \citep{guu2020retrieval}:
\begin{equation}
P(y_i|x) = \sum_z P(y_i|z,x) P(z|x)
\label{eq:mice}
\end{equation}
In Eq.\ref{eq:mice}, $P(z|x)$ represents the likelihood that prompt $z$ is constructed given $x$, and $P(y_i|z,x)$ represents the probability that the LM predicts antecedent $y_i$ when prompted with $z$ and $x$. 

\paragraph{Similarity-based Gating} We compute $P(z|x)$ by summing similarity scores $s(x,u_1),...,s(x,u_d)$ between $x$ and the in-context demonstrations $u_1,...,u_d$ encoded in $z$:\footnote{We also experimented with multiplying the similarity scores and observed similar results.}
\[
P(z|x) \propto \exp{\sum_{i=1}^d s(x,u_i)}
\]
where $s(x, u_i)$ is the cosine similarity between the embeddings of $x$ and $u_i$.  Details of the similarity measures used in our experiments are presented in \S \ref{sec:implementation}.

\noindent
\paragraph{Estimating Antecedent Probabilities}
Computing probabilities $P(y_i|z,x)$ that are comparable across variable-length antecedents is not easy.  Longer sequences will naturally have smaller LM probabilities, suggesting the need for length normalization, or averaging per-token probabilities, neither of which we found to work well.\footnote{Similar to \citet{Zhao2021Calibrate}, we observe that, for a generated antecedent, the first token probabilities vary the most, while probabilities of subsequent tokens are highly deterministic.} 
Therefore, following \citet{Zhao2021Calibrate}, we estimate $P(y_i|z,x)$ using first token probabilities.
Specifically, let $y_{i,0}$ denote the first token of $y_i$. We then have:
\begin{equation}
P(y_i|z,x) \approx \max_j P_j(y_{i,0}|z, x)
\label{eq:pyzx}
\end{equation}
where the max is taken over all mentions of $y_{i}$ found in the $k^d$ prompts. $P_j(y_{i,0}|z,x)$ is the probability of token $y_{i,0}$ being the first token of the $j$th antecedent generated by the language model, when prompted with $z$ and $x$.

\paragraph{Approximation with Sampling}
\label{sec:in_context_sampling}

Approximating $P(y_i|z,x)$ using first-token probabilities in Eq.\ref{eq:pyzx} has a drawback: it disregards the length of the antecedents and treats different antecedents with the same first token as equivalent. As an alternative, we present \micesamp{}: a simple Monte Carlo approximation of $P(y_i|z,x)$ that uses a binary indicator $\mathbbm{1}[y_i \in \mathcal{Y}_{z,x}]$ of whether or not antecedent $y_i$ is in $\mathcal{Y}_{z,x}$ (the set of generated antecedents given prompt $z$ and $x$):
\begin{eqnarray*}
P(y_i|x) & = & \sum_z P(y_i|z,x) P(z|x) \\
    & \approx & \sum_z \mathbbm{1}[y_i \in \mathcal{Y}_{z,x}] P(z|x)
\label{eq:sampling}
\end{eqnarray*}
In \S \ref{sec:experiment}, we show empirically that when using hundreds of prompts, \micesamp{} is actually a better approximation than the LM probability of the first token \citep{Zhao2021Calibrate}. In addition, \micesamp{} is simpler to implement, as it does not require storing output logits and computing the softmax to obtain first token probabilities.

\begin{table}[ht]
\centering
\scalebox{0.75}{
\begin{tabular}{lcccc}
\toprule
 & \textbf{Train (full)} & \textbf{Dev-64} & \textbf{Dev-256} & \textbf{Test} \\ 
\midrule
\# anaphors & 4,766 & 64 & 256 & 898 \\
\# antecedents & 21,673 & 293 & 996 & 4,016 \\
\# documents & 856 & 11 & 52 & 166 \\
\# sentences & 5,833 & 76 & 346 & 1,066 \\
\# tokens & 984,032 & 10,150 & 45,825 & 140,614 \\
\bottomrule
\end{tabular}
}
\caption{Statistics of selected \chemuref{} splits. Details on few-shot train sets are shown in Table \ref{tab:dataset_details} in the Appendix.} 
\label{tab:dataset_summary}
\end{table}

\section{Experimental Setup}
\label{sec:experiment}

We evaluate our approach on \chemuref{}, a corpus of synthetic procedures from chemical patents that are annotated with coreference and anaphora \cite{fang-etal-2021-chemu}. The \chemuref{} annotations contain fine-grained chemistry-specific relations such as \textsc{Transformed}, which indicates the mixture components undergo a chemical transformation, or \textsc{Work-up}, which represents a combination of compounds to isolate or purify a reaction product. In this work, we focus on modeling the general structure of anaphora in scientific protocols and understanding which compounds are combined at each step of the procedure, which does not require making these more fine-grained distinctions. Therefore, we pre-processed the data by collapsing all one-to-many \chemuref{} relations into a single \textsc{Multiple-antecedent} relation.  We remove anaphors comprising compounds described with IUPAC nomenclature \citep{skonieczny2006iupac}, for example {\em ``2,2,6,6-tetramethylpiperidine''}, while retaining nominal anaphors, such as {\em ``the mixture''}, {\em ``the solution''} and {\em ``the reaction''}, which make up 90\% of anaphors in the \chemuref{} corpus.
For train/dev/test splits, we use the original train split for training and development, and the original dev split for evaluation.\footnote{The official \chemuref{} test set is hidden at \url{http://chemu2021.eng.unimelb.edu.au/}.} Following \citet{gao-etal-2021-making}, for each $k$-shot experiment, we sample five different training sets from the full training split using different seeds and report the mean. For model development and ablation studies, we use a small development set of 64 examples (Dev-64) to simulate a true few-shot learning setting. For evaluation on held-out data, we use the full \chemuref{} development set as test data (Test) as well as a sub-sampled version with 256 examples (Dev-256) to control for high GPT-J inference costs. Selected dataset statistics are shown in Table \ref{tab:dataset_summary}.

We use F$_1$ as our evaluation metric. In particular, we compute the micro-F$_1$ between the predicted and gold sets of antecedents of the evaluation data. A predicted and a gold antecedent are considered the same if they are an exact match.

\subsection{Implementation Details}
\label{sec:implementation}

\paragraph{Models} We use GPT-J-6B \cite{gpt-j} as the backbone language model for \mice{} and in-context baselines since it was the largest publicly available autoregressive language model at the time we started this work.\footnote{Larger models, such as OPT-175B \citep{zhang2022opt}, have become available recently. 
}
Compared with other similar language models like GPT-2, GPT-J has a larger maximum sequence length of 2048. 
It is also pre-trained on the Pile \citep{pile}, which covers in-domain data including chemical patents from USPTO\footnote{\url{https://www.uspto.gov/}} and PubMed articles. Similarly, we choose \procbert{} \citep{bai-etal-2021-pre} for the student model in knowledge distillation due to its in-domain pre-training on synthetic procedures.


\paragraph{\mice{}} We ensemble up to 256 prompts, with 2 or 5 in-context demonstrations in the prompt, for all few-shot settings. To calculate the similarity score $s(x, u)$ between the input $x$ and a demonstration example $u$, we use SBERT model \citep{reimers-2019-sentence-bert}\footnote{\url{https://www.sbert.net/docs/pretrained_models.html}} with the \texttt{roberta-large} checkpoint since this model is widely used for measuring text similarity \citep{wang2020cord}.  We also experimented with calibrating the language model to be bias-free by applying the calibration procedure described in \citet{Zhao2021Calibrate}; however, we found that it did not improve performance in the context of anaphora resolution in scientific protocols.\footnote{When prompted with content-free inputs, such as ``N/A'', the language model already generates bias-free answers (e.g. the LM generates "N/A" given "N/A" as a test input) without calibration, suggesting there is no significant bias towards specific answers for anaphora resolution.}


\paragraph{Antecedent Filtering} To further improve the quality of predicted antecedents, we apply several post-processing rules that were developed on the Dev-64 split. 
Namely, we (1) filtered out all predictions that exceed a threshold length of 250 tokens (2) merged the antecedents that are sub-strings to the longest predicted antecedents (e.g. {\em ``CH2CL2''} is merged into {\em ``CH2CL2 (40 mL)'')} and (3) filtered out all the antecedents with probability $P(y_i|z,x)$ in Eq. \ref{eq:pyzx} less than $0.02$ and probability $P(y_i|x)$ in Eq. \ref{eq:mice} less than $0.1$.


\paragraph{Knowledge Distillation} 
To produce a compact model for inference, we perform knowledge distillation \citep{distilbert, jiao-etal-2020-tinybert} via self-training, where the student model is trained on pseudo labels generated \mice{}. 
We experiment with three few-shot settings $k \in \{8, 32, 64\}$. 
In each setting, out of five training runs, we select the \micesamp{}-\{2\} model checkpoint (2 in-context examples) that achieves the median performance on the Dev-256 set as the teacher model.
To train the student model \citep{schick-schutze-2021-just, Lang2022Cotraining},
we randomly sample real unlabeled synthetic protocols from the chemical patent corpus collected in \citet{bai-etal-2021-pre}.
We then run rule-based anaphor detection (Appendix \ref{app:anaphor_detection}) to identify anaphors within the sampled documents.
Subsequently, the teacher model is used to predict antecedents. 
Finally, we train the \procbert{} student model in a two-stage process.
We first fine-tune \procbert{} on $M$ pseudo-labeled examples ($M \in \{50, 100, ..., 2000\}$) for 50 epochs, then further fine-tune it on the $k$ examples with gold labels for 200 epochs. 
For the student model, we frame antecedent resolution as a sequence-labeling problem by transferring span-level antecedent labels into token-level labels with a BIO tagging scheme.
We train the student model for token-level classification, where the input anaphor is marked with two special tokens \texttt{[Ana-start]} and \texttt{[Ana-end]}.

\paragraph{Computational Cost} In addition to performance, we also measure the computational cost of the teacher and student models in knowledge distillation. Concretely, we measure floating point operations (FLOPs) of the two models for both training and inference using the FLOPs-counting code provided in \citet{clark2020electra}\footnote{\url{https://github.com/google-research/electra/blob/master/flops_computation.py}}. Note that, the training of the student model requires $M$ pseudo-labeled examples,
so the training FLOPs of the student model will also include the FLOPs of generating those pseudo labels using the teacher model.

For all GPT-J-based experiments, generation is performed using greedy decoding up to a maximum of 256 tokens on 48GB A40 GPUs.  When using 256 in-context experts, about three anaphors can be resolved per GPU hour.

\begin{table*}[!t]
\centering
\small
\begin{tabular}{lcccccc}
\toprule
\textbf{Model} & \textbf{4-shot} & \textbf{8-shot} & \textbf{16-shot} & \textbf{32-shot} & \textbf{64-shot} & \textbf{full} \\
\midrule
\etoe{} \citep{fang-etal-2021-chemu} & 0.9\textsubscript{1.1} & 10.3\textsubscript{5.3} & 31.6\textsubscript{8.1} & 42.5\textsubscript{5.4} & 51.9\textsubscript{2.3} & 77.4\\
\procbert{} \citep{bai-etal-2021-pre} & 20.3\textsubscript{7.9} & 27.1\textsubscript{3.7} & 36.1\textsubscript{4.1} & 45.3\textsubscript{6.7} & 55.0\textsubscript{4.4} & \textbf{87.7} \\
\textsc{T0-3B} \citep{t0} & 23.5\textsubscript{7.0} & 30.8\textsubscript{4.3} & 42.1\textsubscript{2.3} & 49.5\textsubscript{1.3} & 58.1\textsubscript{3.2} & 83.2\\
\textsc{T5-3B} \citep{t5} & 27.3\textsubscript{3.7} & 33.6\textsubscript{3.9} & 42.5\textsubscript{6.7} & 55.6\textsubscript{6.2} & \bf{61.0}\textsubscript{1.1} & 83.8\\
\kate{} \citep{liu-etal-2022-makes} & 36.2\textsubscript{9.6} & 44.7\textsubscript{5.0} & 46.8\textsubscript{2.9} & 46.4\textsubscript{3.9} & 48.4\textsubscript{1.5} & - \\
\kate{}+ \citep{liu-etal-2022-makes} & 41.9\textsubscript{9.8} & 46.7\textsubscript{10.8} &	48.2\textsubscript{9.0} & 51.4\textsubscript{3.2} & 53.2\textsubscript{1.8}& - \\
\product{} \citep{min-etal-2022-noisy} & 36.7\textsubscript{10.5} & 39.4\textsubscript{8.6} & 41.3\textsubscript{5.3} & 42.4\textsubscript{2.3} & 45.1\textsubscript{2.7} & - \\
\midrule 
\mice{}-\{2\} & 44.4\textsubscript{6.2} & 49.8\textsubscript{5.3} & 52.9\textsubscript{3.8} & 54.6\textsubscript{2.1} &  54.5\textsubscript{3.1} & - \\
\micesamp{}-\{2\} & \bf{44.7}\textsubscript{5.8} & 50.6\textsubscript{5.4} & 55.1\textsubscript{2.9} & 56.8\textsubscript{3.0} & 59.0\textsubscript{3.1} & - \\
\mice{}-\{5\} & 44.5\textsubscript{6.5} & \bf{52.6}\textsubscript{5.0} & 55.8\textsubscript{4.4} & 57.4\textsubscript{1.2} & 58.4\textsubscript{1.9} & - \\
\micesamp{}-\{5\} & 44.4\textsubscript{6.5} & 52.5\textsubscript{5.0} & \textbf{56.8}\textsubscript{4.0} & \textbf{57.5}\textsubscript{3.4} & 59.2\textsubscript{2.1} & - \\
\bottomrule
\end{tabular}

\caption{Results on Dev-256. Bracketed numbers indicate the maximum number of demonstrations per prompt.  For instance, \mice{}-\{2\} represents a mixture of $\min(k^2,256)$ prompted language models where each prompt encodes two training examples.
For each $k$-shot experiment, we report the mean and standard deviation across five different training set samples. 
The {\bf full} \chemuref{} training split contains 4,766 examples.  Results on the test set are presented in Table \ref{tab:test_full}.}
\label{tab:main}
\end{table*}

\subsection{Baselines}

We compare \mice{} to the fine-tuning and in-context learning baselines described below. We use the $k$-shot training set to train the fine-tuning models or select in-context demonstrations for the in-context models. For all baselines, we use the development set Dev-64 for model selection and report the results on the held-out Dev-256 and test sets. Details on baseline implementation can be found in Appendix \ref{app:base_details}.

\paragraph{\etoe{}} \citep{fang-etal-2021-chemu}: We train this end-to-end neural anaphora resolution model developed for chemical procedures with minimal adaptations. \footnote{\url{https://github.com/biaoyanf/ChEMU-Ref}} 
\paragraph{\procbert{}} \citep{bai-etal-2021-pre}: The student model in knowledge distillation (\S\ref{sec:implementation}). For baseline comparison, we fine-tune only on gold data.
\paragraph{T5/T0-3B} \citep{t5,t0}: We fine-tune T5-3B (3 billion parameters) and T0-3B for antecedent resolution using a QA prompt similar to \mice.
\paragraph{\kate{}} \citep{liu-etal-2022-makes}: For this model, each test example is associated with a single prompt constructed using K-nearest neighbors.  For the sentence encoder, we use a pre-trained RoBERTa large model (\texttt{all-roberta-large-v1}) from the SBERT library \citep{reimers-2019-sentence-bert}.

\paragraph{\kate{}+} Given a single prompt constructed by \kate{}, we sample 256 antecedent lists using nucleus sampling \citep{nucleusSampling} and ensemble the predictions in a similar manner to \mice{}. This baseline ensembles the same number of predictions as \mice{}, with the difference being they are all sampled from the language model when conditioned on a single prompt.
    
\paragraph{\product{}} \citep{min-etal-2022-noisy}: We adapt this ensemble-based demonstration method for text classification by first obtaining the output probabilities $k$ times (one training example per prompt) and then computing the conditional probability of the antecedents given the input by taking the product of the aforementioned probabilities (i.e. $P(y|x) = \prod_z P(y|z,x)$).
    
    

\section{Results and Analysis}
\label{sec:results}


\begin{table}[!t]
\centering
\scalebox{0.65}{
\begin{tabular}{lcccc}
\toprule 
\textbf{Model} & \textbf{8-shot} & \textbf{32-shot} & \textbf{64-shot} & \textbf{full} \\ 
 \midrule
\etoe{} \citep{fang-etal-2021-chemu} & 9.9 & 42.3 & 51.8 & 75.8 \\ 
\textsc{ProcBERT} \citep{bai-etal-2021-pre} & 30.8 & 41.9 & 57.6 & \textbf{78.3} \\ 
T5-3B \citep{t5} & 33.9 & 52.0 & 58.6 & 72.6 \\
\kate{} \citep{liu-etal-2022-makes} & 37.9 & 38.6 & 42.3 & - \\ 
\textsc{Product} \citep{min-etal-2022-noisy} & 34.3 & 38.5 & 49.0 & - \\
 \midrule
\textsc{Product-\{2\}} & 45.6 & 51.1 & 51.3 & - \\ 
\mice{}-\{2\} & 48.1 & 52.6 & 53.9 & - \\ 
\micesamp{}-\{2\} & 48.5 & 53.9 & 55.7 & - \\ 
Know. Distill. (2000 exam.) & \textbf{56.2} & \textbf{59.4} & \textbf{64.6} & - \\ 
 \bottomrule
\end{tabular}
}
\caption{F$_1$ on the full test set.
For the knowledge distillation model, we picked the median trial from \micesamp{}-\{2\} of Table \ref{tab:main} to run experiment on, due to the expensive inference cost on large amount of unlabeled data. For others, we averaged over five training samples. 
}
\label{tab:test_full}
\end{table}

\begin{figure}[!h]
    \centering
    \includegraphics[width=0.50\textwidth]{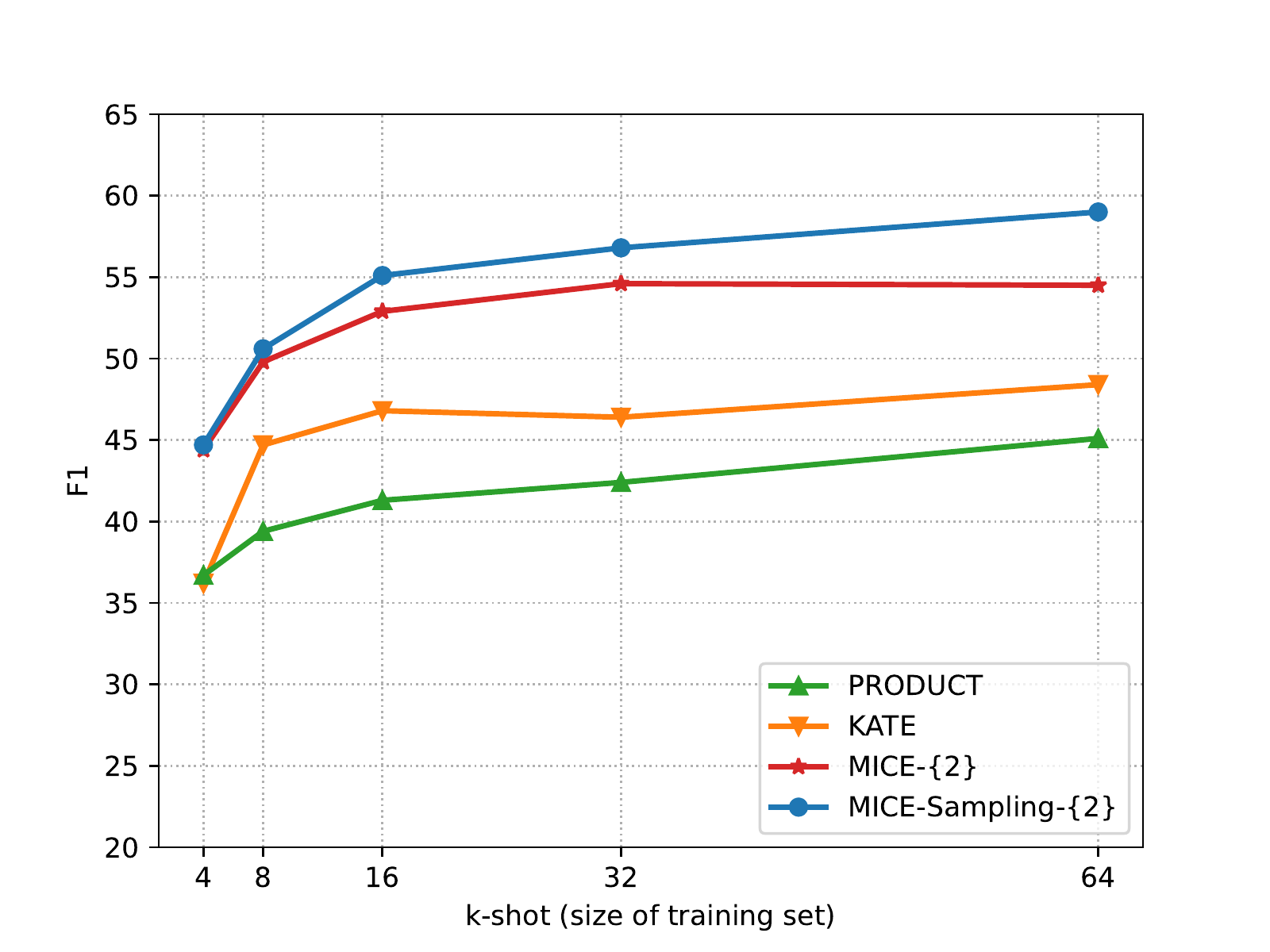}
    \caption{Mean F$_1$ of in-context learning models vs. training set size $k$, on Dev-256.}
    \label{fig:mice_vs_single_wrt_K}
\end{figure}

Results on Dev-256 are presented in Table \ref{tab:main}. We observe that variations of \mice{} either outperform (4-shot, 8-shot, 16-shot, 32-shot) or are competitive (64-shot) with the baselines. Furthermore, we found that while T5-3B works well for 64-shot, it still trails behind \mice{} and \micesamp{} (and even \kate{}) when using fewer labeled data.\footnote{We also experimented with T-few \citep{liu2020tfew}, a performant T0-based parameter-efficient fine-tuning approach.
With suggested hyper-parameters, T-few performs worse than the full-model fine-tuning of T0-3B on anaphora resolution, so we leave the further exploration of T-few to future work.}
Additionally, we observe that \mice{} outperforms other in-context learning models that use a single most performant prompt (\kate{}) or a single training example per prompt (\product{}). This is highlighted in Figure \ref{fig:mice_vs_single_wrt_K}, where we plot the performance of various in-context learning models as a function of training set size $k$. 

Results on the full test set (Table \ref{tab:test_full}) shows similar trends. We also experiment with extending the \product{} baseline to include two in-context demonstrations per prompt (\product{}-\{2\}). While increasing the number of in-context demonstrations in \product{} also boosts the performance for all $k$-shot, \mice{} and \micesamp{} still outperform \product{}-\{2\}.

\paragraph{\kate{}+ vs. \mice{}} We highlight the results of \kate{}+ in comparison to \kate{} and \mice{}. \kate{}+ performs somewhat better than \kate{}, likely due to state-of-the-art text generation methods (e.g. nucleus sampling) as well as ensembling independently-sampled sequences conditioned on a single prompt. However, \mice{} significantly outperforms \kate{}+, indicating the importance of ensembling prompts with different demonstrations and permutations.

\begin{table}[!t]
\centering
\resizebox{0.48\textwidth}{!}{
\begin{tabular}{llccccc}
\toprule 
\multirow{2}{*}{\textbf{Model}} & & \multicolumn{3}{c}{\textbf{F\textsubscript{1}}} & \multirow{2}{*}{\textbf{Train FLOPs}} & \textbf{Infer FLOPs} \\ 
& & 8-shot & 32-shot & 64-shot &  & (1000 exam.) \\
 \midrule
\multicolumn{2}{l}{\procbert{}} & 28.0 & 41.1 & 56.7 & 3.2e15 & 1.2e14 \\ 
\midrule
\multicolumn{2}{l}{\textbf{Teacher}} & 54.1 & 55.7 & 59.9 & 0 & 1.9e17 \\
\multicolumn{6}{l}{\textbf{Student} (\# pseudo labels)} \\
\hspace{0.5mm} - 50   & & 36.0 & 46.2 & 58.4 & 1.4e16 & 1.2e14 \\
\hspace{0.5mm} - 100  & & 43.5 & 49.9 & 62.2 & 2.4e16 & 1.2e14 \\
\hspace{0.5mm} - 200  & & 42.1 & 52.3 & 60.4 & 4.5e16 & 1.2e14 \\
\hspace{0.5mm} - 500  & & 49.1 & 52.2 & 63.3 & 1.1e17 & 1.2e14 \\
\hspace{0.5mm} - 1000 & & 50.1 & 51.8 & 64.1 & 2.1e17 & 1.2e14 \\
\hspace{0.5mm} - 2000 & & \textbf{54.3} & \textbf{55.8} & \textbf{64.3} & 4.1e17 & 1.2e14 \\
 \bottomrule
\end{tabular}
}
\caption{Dev-256 F\textsubscript{1} and training/inference FLOPs of the teacher model (\micesamp{}) and the student model in knowledge distillation. 
The training FLOPs of the student model under different few-shot settings are almost the same.
With 2000 pseudo-labeled examples, the student model (110M parameters) can match or outperform the teacher model in terms of F\textsubscript{1}.
Although inference FLOPs of the teacher model are around 1500 times the FLOPs of the student model, 
considering the cost of training the student model to match the teacher model's performance, the teacher model is actually more cost effective than the student model for inference when the number of inference examples is low, e.g., less than 2100 for 32-shot.  For rows where the number of FLOPs varies depending on the number of training examples, we show the maximum.
}
\label{tab:knowledge_distill}
\end{table}

\paragraph{Knowledge Distillation} 
The impact of knowledge distillation is explored in Table \ref{tab:knowledge_distill}.
We observe that with only 50 pseudo-labeled examples, the student model outperforms the \procbert{} baseline by 8.0 F\textsubscript{1} for 8-shot, 5.1 F\textsubscript{1} for 32-shot and 1.7 F\textsubscript{1} for 64-shot, showing the effectiveness of distillation. The performance of the student model improves as the number of pseudo-labeled examples increases. 
With 2000 pseudo-labeled examples, the inference-efficient student model (110M parameters) outperforms the teacher model \micesamp{}.  We hypothesize that the student model outperforms the teacher because after training on pseudo-labels generated by MICE, the student model is further fine-tuned on the $k$ available gold-labeled examples, as described in \S \ref{sec:implementation}.  This approach combines the benefits of in-context learning with fine-tuning achieving better performance than either in isolation.
In-context learning does not incur any training costs, but the number of FLOPs required by \micesamp{} for inference is roughly 1,500 times the computational cost of the student model.


\begin{figure}[!t]
    \centering
    \includegraphics[scale=0.50]{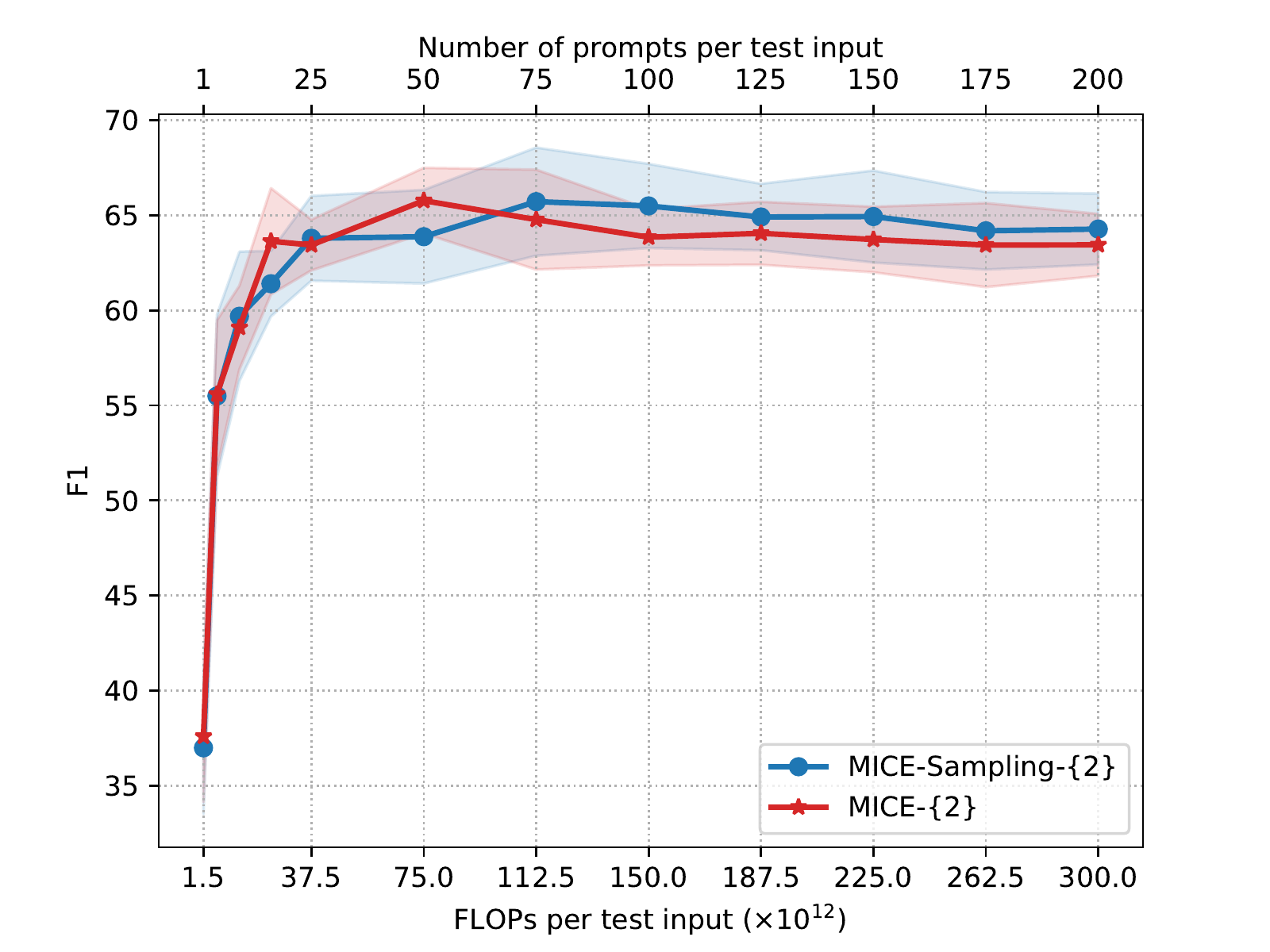}
    \caption{Mean F$_1$ ($\pm$ one std. dev.) vs. the number of prompts used in {\sc Mice}. 
    Performance plateaus after the number of experts is increased to 75, suggesting an appropriate choice for the minimum number of prompts in this context. The number of prompts corresponds to the number of inference passes, which we measure using FLOPs per test example (\S \ref{sec:implementation}).}
    \label{fig:f1_vs_numPrompts}
\end{figure}

\paragraph{Number of mixture prompts vs. Performance} We explore the effect of varying the number of mixture prompts in Figure \ref{fig:f1_vs_numPrompts}. While adding more prompts leads to rapid F$_1$ improvement, the performance plateaus after about 75 prompts. This suggests an appropriate minimum number of experts to use in \mice{}. 

\begin{figure}[!t]
    \centering
    \includegraphics[width=0.50\textwidth]{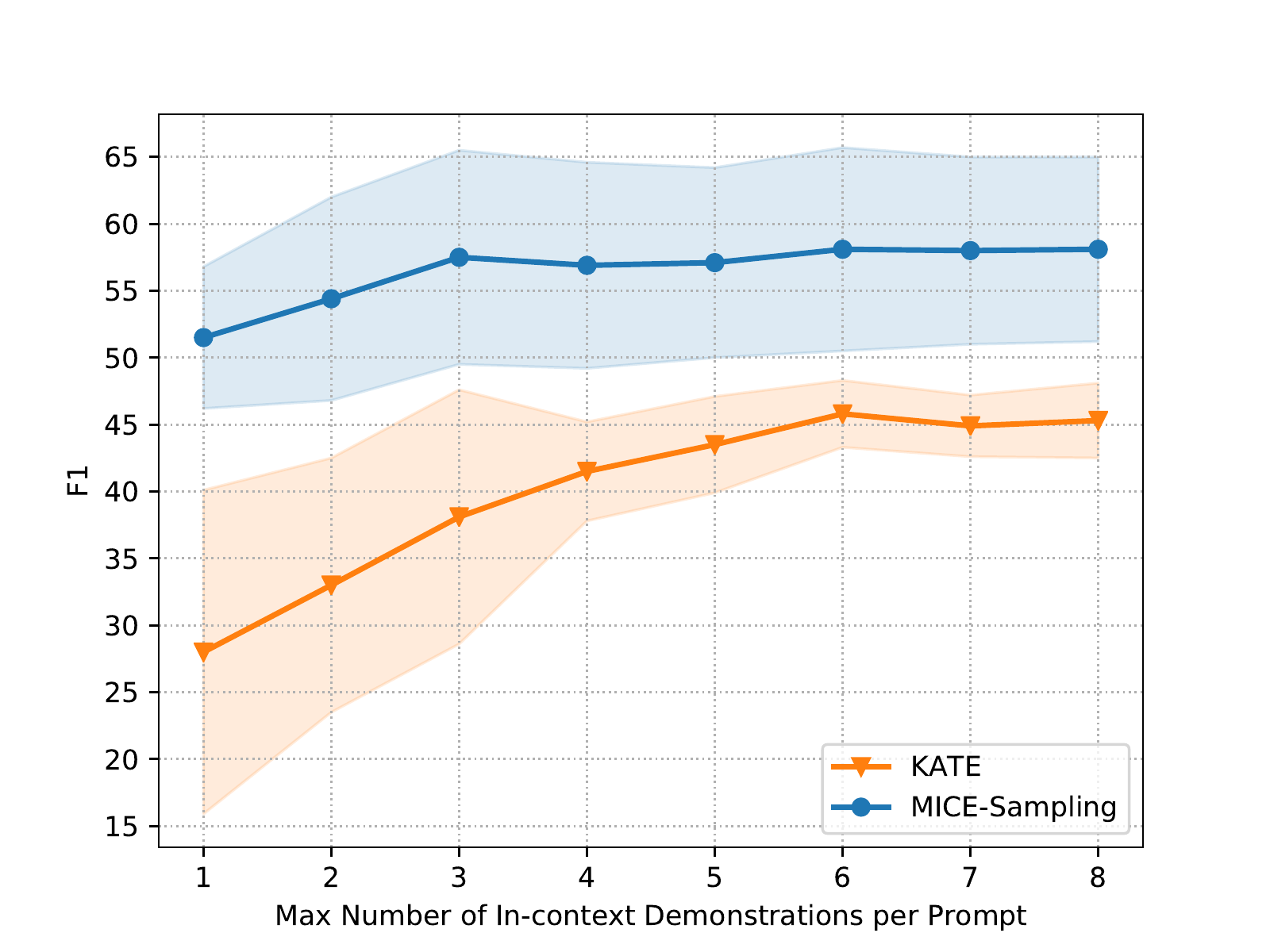}
    \caption{Mean F$_1$ ($\pm$ one std. dev.) vs. max number of in-context demonstrations per prompt.
    We observe a large performance gap between \micesamp{}{} and \kate{}.}
    \label{fig:f1_vs_numinContext_k08}
\end{figure}

\paragraph{Number of in-context demonstrations vs. Performance} Figure \ref{fig:f1_vs_numinContext_k08} shows the effect of varying the number of in-context demonstrations per prompt, for mixture models (\micesamp{}) and single-prompted models (\kate{}). As the number of in-context demonstration per prompt increases, \kate{} F$_1$ steadily increases. This effect is less pronounced in \mice{}, where adding more than three in-context demonstrations per prompt does not significantly improve performance. However, there is still a large performance gap between \mice{} and \kate{}, regardless of the number of in-context demonstrations.


\section{Related Work}
\label{sec:related}



\paragraph{Ensembling prompts} 
Recent work has explored different strategies for ensembling LM predictions. 
\citet{lester2021power} and \citet{zhao2022lmturk} aggregate predictions from multiple prompts via either majority voting or weighting on development set performances. \citet{jiang2020can} and \citet{qin2021learning} ensemble a small number of prompt templates or soft prompts, then learn their weights from a large training set.
\citet{asai2022attentional} combines soft prompts pre-trained on multiple source tasks with fine-tuned attention weights to generate prompts for a target task.
In contrast to all the above work, \mice{} generates a large number of prompts from a small number of training examples and combines their predictions using mixtures of experts with similarity-based gating.

Similar to our work, \citet{min-etal-2022-noisy} shows that ensembling the predictions of $k$ one-shot prompts by multiplying their LM probabilities is better than concatenating all $k$ examples as one prompt, which we use as a baseline (\product{}). 
Concurrently, \citet{Lang2022Cotraining} demonstrates the advantage of ensembling $k$ one-shot prompts via co-training the ensembling parameters using a smaller model. 
However, they train the ensembling parameters with an MLP layer and thus require all training examples to share the same label space, which is not applicable to anaphora resolution.
Importantly, both \citet{min-etal-2022-noisy} and \citet{Lang2022Cotraining} only consider a small number of prompts, whereas we demonstrate the benefits of generating and combining hundreds of prompts as a mixture-of-experts.


\paragraph{Few-shot anaphora resolution} 
There is very little prior work on few-shot coreference or anaphora resolution.
Prior work \citep{perez2021true, min2022metaicl} has applied in-context learning to the benchmark datasets Winograd Schema Challenge (WSC) \citep{levesque2012winograd} and WinoGrande \citep{sakaguchi2019winogrande}. Both datasets are intended to be an alternative to the Turing Test. As such, they contain short, syntactically simple sentences, and require a different type of world knowledge to resolve references than those seen in scientific protocols. \citet{ClinicalLM} demonstrates the effectiveness of zero-shot prompting for coreference resolution on clinical data. They introduce task-specific programs (called \textit{resolvers}) that map language model outputs to discrete label space, which can be used in conjunction with \mice{} for further improvements on few-shot coreference resolution. 



\paragraph{Mixture-of-Experts in Language Modelling}
Mixture-of-Experts (MoE) based language models have been shown to improve performance and efficiency across a variety of NLP tasks \cite{SparseMoE, GLAM, gururangan-etal-2022-demix}. These models were pre-trained on a mixture of datasets with different domains, either via learning the gating weights at a token level \cite{SparseMoE, GLAM} or document level \cite{gururangan-etal-2022-demix}. Unlike prior work, \mice{} combines the predictions of many in-context experts, each of which encodes a permutation of demonstrations drawn from a handful of training examples.  The gating weights in MICE are computed based on similarity scores between test input and in-context examples.

\section{Conclusion}
\label{sec:conclu}

In this paper, we propose and demonstrate the effectiveness of Mixtures of In-Context Experts for few-shot anaphora resolution in chemical synthesis protocols. \mice{} generates a large number of in-context experts (prompts) from a few training examples, where each expert consists of randomly permuted demonstrations. Predictions made by these experts are combined in a mixture model with a similarity-based gating function.
Our experiments show that \mice{} significantly improves the performance of in-context learning for anaphora resolution in scientific protocols, using just a handful of training examples.
We further demonstrate that knowledge distillation can dramatically reduce the costs of inference while maintaining performance, which increases \mice{}'s potential applicability in working systems.


\section*{Limitations}
\label{sec:limitations}
A challenging bottleneck for \mice{} is its expensive inference cost: despite the promising few-shot capabilities, the inference cost is even more expensive when using a language model with over 6 billion parameters.
Although we show that knowledge distillation can address this problem, future work may further investigate how to improve smaller models' in-context capabilities.



\section*{Acknowledgements}
This material is based upon work supported by the Defense Advanced Research Projects Agency (DARPA) under Contract No. HR001119C0108, in addition to the NSF (IIS-2052498) and IARPA via the BETTER program (2019-19051600004).
The views, opinions, and/or findings expressed are those of the author(s) and should not be interpreted as representing the official views or policies of the Department of Defense, IARPA or the U.S. Government.  This work is approved for Public Release, Distribution Unlimited.

\bibliography{emnlp2022}
\bibliographystyle{acl_natbib}

\clearpage

\appendix

\section{Anaphor Detection}
\label{app:anaphor_detection}

We develop a rule-based system using simple string matching patterns. We developed the system using the full train set, optimized on the Dev-64 development set, and evaluated on Dev-256 and Test set. 
Anaphora in synthetic protocols are expressed using relatively consistent and easy to identify expressions, for example: \textit{"the mixture", "the reaction solution"}, and \textit{"the resulting solution"}, making them relatively easy to identify using a simple approach using regular expressions. 
As a result, our system can effectively identify most of these spans, achieving high F$_1$ scores when evaluated on different evaluation splits (Table \ref{tab:anaphor_detection}). An example of a pattern that matches all anaphors in Figure \ref{fig:approach} test input is shown here: \url{https://regex101.com/r/pImmBT/1}

\begin{table}[!ht]
\centering
\scalebox{0.9}{
\begin{tabular}{lccc}
\toprule
 & \textbf{P} & \textbf{R} & \bf{F$_1$} \\
\midrule
Dev-64 & 98.5 & 100 & 99.2 \\
Dev-256 & 97.6 & 95.7 & 96.6 \\
Test & 95.6 & 85.5 & 90.3 \\
\bottomrule
\end{tabular}
}
\caption{Results of our rule-based anaphor detection system on different evaluation splits.}
\label{tab:anaphor_detection}
\end{table}

\section{\mice{} Implementation Details}
\label{app:prompt_hyperparameters}

We present the hyperparameters for \mice{} (as well as other in-context learning baselines) in Table \ref{tab:mice_hyperpara}. We also experimented with different prompt ordering of in-context demonstrations \citet{liu-etal-2022-makes} and contextual calibration \citet{Zhao2021Calibrate}. All ablation experiments were evaluated on Dev-64 development set.

\paragraph{Ordering of In-context Demonstrations} Since different prompt ordering can play a significant role in the performance, we measure the effect of different in-context ordering on \kate{}. Given a test input, we first select the most similar examples using based on the cosine similarity between the test input and the example embeddings (encoded using pre-trained RoBERTa large model). We then consider the following ordering:
\begin{itemize}
    \item \textsc{Ascend}: Sorting the order of in-context demonstrations from least to most similar examples (i.e. most similar example closest to test input)
    \item \textsc{Descend}: Reverse of \textsc{Ascend}
    \item \textsc{Mixed}: Random shuffle of in-context demonstrations
\end{itemize}
The results on Table \ref{tab:in_context_ordering} show there is not much difference between different orderings. This result is in line with the findings in \citet{lu-etal-2022-fantastically} that prompt ordering is dataset-dependent. As such, we select \textsc{Ascend} as the de facto ordering for all our in-context learning experiments.

\begin{table*}[t]
\centering
\scalebox{0.9}{
\begin{tabular}{lccccc}
\toprule
& \textbf{4-shot} & \textbf{8-shot} & \textbf{16-shot} & \textbf{32-shot} & \textbf{64-shot} \\
\midrule
\textsc{Ascend} & 34.3\textsubscript{4.9} & 45.3\textsubscript{2.8} & \textbf{46.8}\textsubscript{5.1} & \textbf{46.8}\textsubscript{6.6} & \textbf{46.5}\textsubscript{5.9} \\
\textsc{Descend} & \textbf{34.8}\textsubscript{4.8} & \textbf{46.2}\textsubscript{1.3} & 45.1\textsubscript{4.8} & 44.6\textsubscript{6.6} & 46.5\textsubscript{4.7} \\
\textsc{Mixed} & 34.1\textsubscript{7.6} & 43.9\textsubscript{3.3} & 46.1\textsubscript{4.7} & 46.2\textsubscript{4.3} & 46.3\textsubscript{4.1} \\
\bottomrule
\end{tabular}
}
\caption{Order of in-context demonstration results.}
\label{tab:in_context_ordering}
\end{table*}

\paragraph{Contextual Calibration} \citet{Zhao2021Calibrate} shows that prompted LMs can be biased towards specific outputs, regardless of the test input. We thus experimented with the calibration procedure for generation tasks described in Section 5 of \citet{Zhao2021Calibrate}, also referencing their code.\footnote{\url{https://github.com/tonyzhaozh/few-shot-learning}} Specifically, we computed calibration parameters using the output probabilities of the content-free prompt (e.g. prompt with "N/A" test input), and used these parameters to update output probabilities when prompted with actual test inputs. However, we found that applying this calibration procedure does not improve performance. We further observe that when prompted with content-free inputs, the language model already generates bias-free answers (e.g. generates "N/A" given "N/A" test input) without calibration. This suggests there is no significant bias towards specific answers for the case of split-antecedent resolution, whereas current calibration techniques are designed for cases where the LM generates biased answers given content-free inputs, (e.g. generates a label given "N/A" test input). 

\begin{table*}[!t]
\centering
\scalebox{0.8}{
\begin{tabular}{lccccc}
\toprule 
\multirow{2}{*}{\textbf{Hyperparameter}} & \multirow{2}{*}{\textbf{\kate{}}} & \multirow{2}{*}{\textbf{\kate{}}+} & \multirow{2}{*}{\textbf{\product{}}} & \textbf{\mice{}-\{2\}} & \textbf{\mice{}-\{5\}} \\
 & &  &  &  \textbf{\micesamp{}-\{2\}} &  \textbf{\micesamp{}-\{5\}}\\ 
 \midrule
\# prompts per example & 1  & 1 & $k$ & $\min(256, k^2)$ & $\min(256, k^2)$ \\
Max \# in-context per prompt & $\min(k, d_x)$ & $\min(k, d_x)$  & 1 & 2 & 5 \\
Order of in-context & \textsc{Ascend} & \textsc{Ascend} & - & \textsc{Ascend} & \textsc{Ascend} \\
$P(y_i|z,x)$ threshold & 0.0  & 0.02 & 0.02 & 0.02 & 0.02 \\ 
$P(y_i|x)$ threshold & 0.0 & 0.1 & 0.1 & 0.1 & 0.1 \\
top-$k$  & - & 50 & - & - & - \\
top-$p$  & - & 0.95 & - & - & - \\
 \bottomrule
\end{tabular}
}
\caption{Hyperparameters for in-context learning models. $k$ is the size of training set ($k$-shot). $d_x$ depends on the length of test input $x$ and retrieved in-context examples (Figure \ref{fig:dev64_maxInContextNum}). \textsc{Ascend} denotes sorting the order of in-context demonstrations from least to most similar examples (i.e. most similar example closest to test input). Top-$k$ and top-$p$ are hyperparameters for nucleus sampling used for \kate{}+.}
\label{tab:mice_hyperpara}
\end{table*}




\section{Baseline Implementation Details}
\label{app:base_details}
For \etoe{} \citep{fang-etal-2021-chemu}, we run the TensorFlow code provided in the original paper (following the same hyperparameters) to get the model performance under our few-shot settings.
Other baselines, including \procbert{}, T5-3B, and T0-3B, are implemented using Huggingface Transformers \citep{wolf-etal-2020-transformers}. The hyperparameters used for these baselines are shown in Table \ref{tab:base_hyperpara}.

\begin{table*}[!t]
\centering
\scalebox{0.8}{
\begin{tabular}{lcc}
\toprule 
\textbf{Hyperparameter}&  \textbf{\procbert{}} & \textbf{T5-3B} / \textbf{T0-3B} \\ 
 \midrule
\# epochs& 200 & 20 \\ 
batch size & 32 & 4\\
learning rate & 2e-5 & 1e-4 \\
max. encoder length & 512 & 512 \\
max. decoder length & - & 256 \\
optimizer & AdamW & AdamW \\
 \bottomrule
\end{tabular}
}
\caption{Hyperparameters for \procbert{}, T5-3B, and T0-3B. 
}
\label{tab:base_hyperpara}
\end{table*}


\begin{table*}[!t]
\centering
\scalebox{0.9}{
\begin{tabular}{lccccccccc}
\toprule
\textbf{Data Split} & \textbf{4-shot} & \textbf{8-shot} & \textbf{16-shot} & \textbf{32-shot} & \textbf{64-shot} & \textbf{Train (full)} & \textbf{Dev-64} & \textbf{Dev-256} & \textbf{Test} \\ 
\midrule
\# anaphors & 4 & 8 & 16 & 32 & 64 & 4,766 & 64 & 256 & 898 \\
\# antecedents & 13.4 & 24.2 & 61.2 & 132.8  & 266.0 & 21,673 & 293 & 996 & 4,016 \\
\# documents & 2.2 & 3.8 & 6.0  & 9.6 & 17.6 & 856 & 11 & 52 & 166 \\
\# sentences & 8.0 & 15.2 & 27.0 & 50.4 & 92.2 & 5,833 & 76 & 346 & 1,066 \\
\# tokens & 445.8 & 939.6 & 2,257.6 & 4,665.2 & 9,775.8 & 984,032 & 10,150 & 45,825 & 140,614 \\
\bottomrule
\end{tabular}
}
\caption{Dataset splits details. The $k$-shot dataset statistics are averaged over five random samples.}
\label{tab:dataset_details}
\end{table*}

\end{document}